\documentclass[10pt,twocolumn,letterpaper]{article}

\usepackage[pagenumbers]{cvpr} 

\usepackage{graphicx}
\usepackage{amsmath}
\usepackage{amssymb}
\usepackage{booktabs}
\usepackage{placeins}

\usepackage{colortbl}
\usepackage{tabulary}
\usepackage[table]{xcolor}
\usepackage{color} 
\usepackage{xcolor}
\usepackage{multirow}
\usepackage{bbding}

\usepackage[pagebackref,breaklinks,colorlinks]{hyperref}

\usepackage[capitalize]{cleveref}
\crefname{section}{Sec.}{Secs.}
\Crefname{section}{Section}{Sections}
\Crefname{table}{Table}{Tables}
\crefname{table}{Tab.}{Tabs.}

\def\cvprPaperID{2454} 
\def\confName{CVPR}
\def\confYear{2023}

\makeatletter
\newbox\abstract@box
\renewenvironment{abstract}
  {\global\setbox\abstract@box=\vbox\bgroup
     \hsize=\textwidth\linewidth=\textwidth
    \small
    \begin{center}%
    {\bfseries \abstractname\vspace{-.5em}\vspace{\z@}}%
    \end{center}%
    \quotation}
  {\endquotation\egroup}
\expandafter\def\expandafter\@maketitle\expandafter{\@maketitle
  \ifvoid\abstract@box\else\unvbox\abstract@box\if@twocolumn\vskip1.5em\fi\fi}
\makeatother

\begin{document}

\title{BEV-SAN: Accurate BEV 3D Object Detection via Slice Attention Networks}

\author{Xiaowei Chi\textsuperscript{\rm 1,4}\thanks{Equal contribution}, 
Jiaming Liu\textsuperscript{\rm 1*},
Ming Lu\textsuperscript{\rm 1*},
Rongyu Zhang\textsuperscript{\rm 2},\\
Zhaoqing Wang\textsuperscript{\rm 3},
Yandong Guo\textsuperscript{\rm 5},
Shanghang Zhang\textsuperscript{\rm 1}\thanks{Corresponding author: shzhang.pku@gmail.com}\\
\textsuperscript{\rm 1}Peking University, \textsuperscript{\rm 2}The Chinese University of Hong Kong, Shenzhen,
\textsuperscript{\rm 3} The University of Sydney\\
\textsuperscript{\rm 4}The Chinese University of Hong Kong,
\textsuperscript{\rm 5}Beijing University of Posts and Telecommunications\\ 
}

\begin{abstract}
Bird's-Eye-View (BEV) 3D Object Detection is a crucial multi-view technique for autonomous driving systems. Recently, plenty of works are proposed, following a similar paradigm consisting of three essential components, i.e., camera feature extraction, BEV feature construction, and task heads. Among the three components, BEV feature construction is BEV-specific compared with 2D tasks. Existing methods aggregate the multi-view camera features to the flattened grid in order to construct the BEV feature. However, flattening the BEV space along the height dimension fails to emphasize the informative features of different heights. For example, the barrier is located at a low height while the truck is located at a high height. In this paper, we propose a novel method named BEV Slice Attention Network (BEV-SAN) for exploiting the intrinsic characteristics of different heights. Instead of flattening the BEV space, we first sample along the height dimension to build the global and local BEV slices. Then, the features of BEV slices are aggregated from the camera features and merged by the attention mechanism. Finally, we fuse the merged local and global BEV features by a transformer to generate the final feature map for task heads. The purpose of local BEV slices is to emphasize informative heights. In order to find them, we further propose a LiDAR-guided sampling strategy to leverage the statistical distribution of LiDAR to determine the heights of local slices. Compared with uniform sampling, LiDAR-guided sampling can determine more informative heights. We conduct detailed experiments to demonstrate the effectiveness of BEV-SAN. Code will be released.
\end{abstract}
\maketitle
\section{Introduction}
\label{sec:intro}

Object detection is an essential computer vision task, which has wide applications in security, robotics, autonomous driving, etc. With the development of Deep Neural Networks (DNNs), a huge amount of methods are proposed for 2D \cite{girshick2014rich,girshick2015fast,he2017mask,liu2021swin,ren2015faster,redmon2016you} and 3D \cite{chen2017multi,shi2020pv,yang20203dssd,qi2019deep} object detection. As there are too many methods, we focus our introduction on the cutting-edge multi-view camera-based 3D object detection, which has gained increasing attention from the community. The Bird's-Eye-View (BEV) is a unified representation of the surrounding scene and is suitable for autonomous driving tasks. Therefore, plenty of 3D object detection methods \cite{wang2022detr3d,huang2021bevdet,liu2022petr,li2022bevformer,li2022bevdepth,xu2022cobevt,li2022bevstereo,huang2022bevdet4d,chen2022efficient} are proposed for multi-view BEV perception recently.

\begin{figure}[t]
	\includegraphics[scale=0.5]{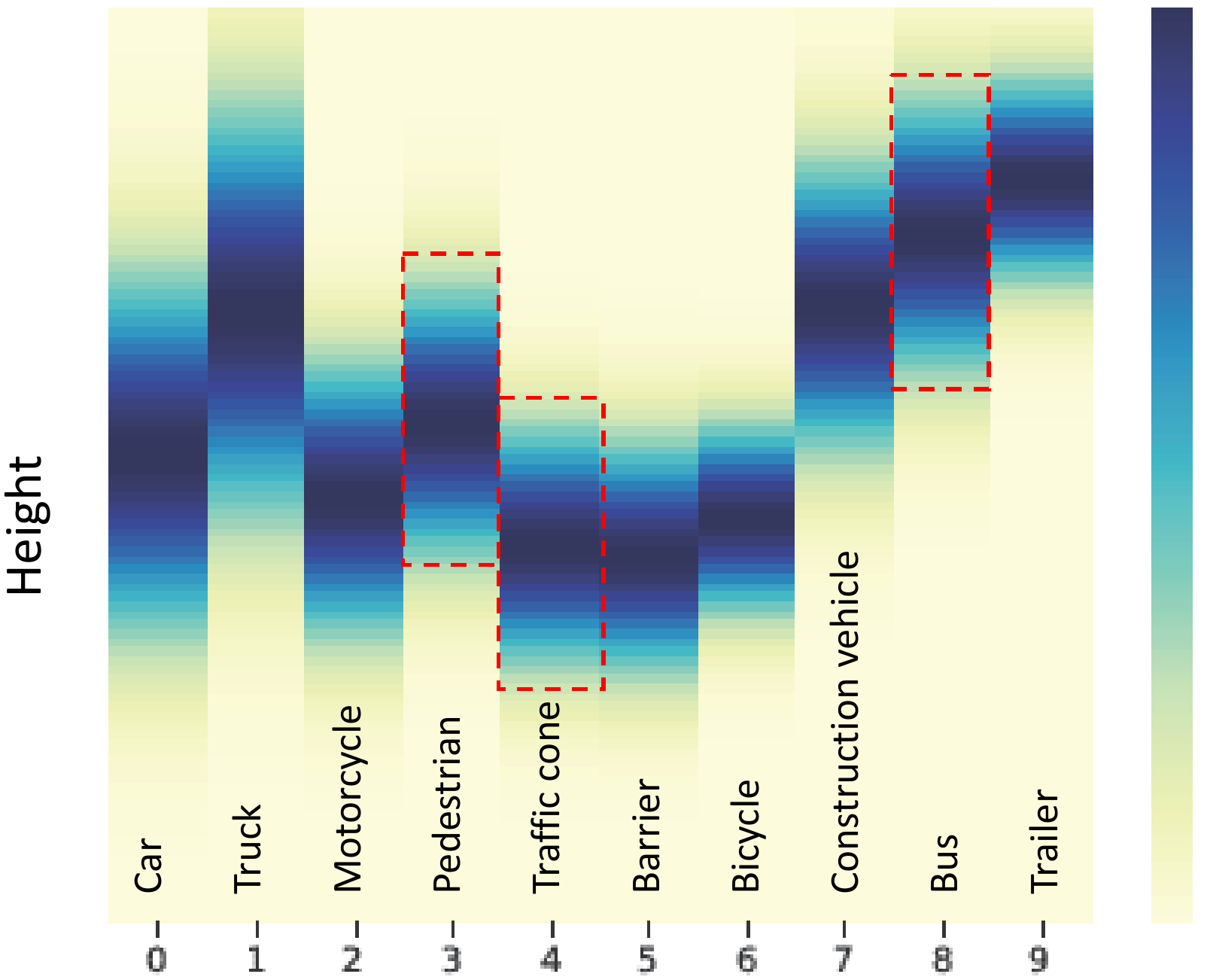}
	\centering
	\caption{The statistics of 3D bounding boxes along the height dimension.}
 \label{fig:heat_hight_cropped}
\end{figure}

Although the model architectures of those methods are different, they commonly follow a similar paradigm consisting of three essential components including camera feature extraction, BEV feature extraction, and task heads. Among the three components, BEV feature construction is BEV-specific compared with 2D tasks. \cite{li2022bevformer} presents a new framework that learns a unified BEV representation with spatio-temporal transformers. They first lift each query on the flattened BEV grid to a pillar-like query and then project the sampled 3D points to 2D views. The extracted features of hit views are weighted and summed as the output of spatial cross-attention. \cite{li2022bevdepth} first predicts the depth for RGB input and projects the image features to frustum space. Then they sum up the frustum features that fall into the same flatted BEV grid. Both methods have pros and cons, while they all flatten the BEV space along the height dimension.

Motivated by the fact that different object classes locate at different heights. For instance, barrier is located at a low height while the truck is located at a high height. Flattening the BEV space along the height dimension fails to exploit the benefit of different heights. In this paper, we propose a novel method named BEV Slice Attention Network (BEV-SAN) to explore the intrinsic properties of different heights. We first sample along the height dimension to build the global and local BEV slices, which are represented as the upper and lower bounds of BEV slice height. The global slices are similar to former works \cite{li2022bevformer,li2022bevdepth}, which aim at covering the large height range of BEV space, while the local BEV slices aim at emphasizing informative heights. We aggregate the features from multi-view cameras to construct the features of global and local BEV slices. To merge the global and local slices, we first use the height attention mechanism to fuse the global and local slices separately. Then we adopt a transformer to fuse the merged global and local features. The final fused feature map is used for task-specific heads. In this paper, we mainly conduct the evaluation of BEV-SAN on 3D object detection. It is to be noted that our method can also be used in other BEV perception tasks such as map segmentation and planning.

In order to improve the performance, we further propose a LiDAR-guided sampling strategy to leverage the statistical distribution of LiDAR to determine the optimal heights of local slices. We project the LiDAR points to the BEV space and calculate the histogram along the height dimension. According to the histogram, we can sample the upper and lower height bounds of local slices. Compared with uniform sampling or random sampling, our strategy can choose informative ranges for BEV perception. We want to point out that we only use LiDAR data to build the local BEV slices. Our contributions can be concluded as follows:

\begin{itemize}
\item We propose a novel method named BEV Slice Attention Network (BEV-SAN) that exploits the features of different heights in BEV space, achieving an accurate performance of BEV 3D object detection.

\item We present a LiDAR-guided sampling strategy to determine the optimal heights of local slices, resulting in informative ranges for BEV perception.

\item We conduct detailed experiments to demonstrate the effectiveness of our method. Our method can also be applied to other BEV perception tasks like map segmentation and planning.
\end{itemize}

\section{Relate work}

{\bf Monocular 3D object detection} Monocular 3D object detection is a useful but challenging technique in autonomous driving since it needs to predict the 3D bounding boxes from a single 2D image. Deep3DBox \cite{mousavian20173d} firstly regresses relatively stable 3D bounding box properties using DNNs and combines them with geometric constraints to generate the final results. M3D-RPN \cite{brazil2019m3d} designs depth-aware convolutional layers and 3D region proposal network, significantly improving the performance of monocular 3D object detection. SMOKE \cite{liu2020smoke} predicts a 3D bounding box for each detected 2D object by combining a single keypoint estimate with regressed 3D variables. FCOS3D \cite{wang2021fcos3d} proposes a one-stage framework that predicts the decoupled 2D and 3D attributes for 3D targets. MonoDLE \cite{ma2021delving} quantifies the impact introduced by each sub-task of monocular 3D object detection and proposes three strategies to reduce the localization error. PGD \cite{wang2022probabilistic} constructs geometric relation graphs across predicted objects and uses the graph to improve the depth estimation for monocular 3D object detection. MonoPair \cite{chen2020monopair} improves monocular 3D object detection by considering the relationship of paired samples. RTM3D \cite{li2020rtm3d} predicts the nine perspective key points in 3D space and recovers the dimension, location, and orientation from the nine key points. MonoFlex \cite{zhang2021objects} proposes a flexible framework that explicitly decouples the truncated objects and adaptively combines multiple approaches for object depth estimation. GUP-Net \cite{lu2021geometry} proposes to tackle the error amplification problem introduced by the projection process. MonoDETR \cite{zhang2022monodetr} introduces a novel framework using a depth-guided transformer and achieves state-of-the-art performance on benchmarks.

\begin{figure*}[t]
\centering
\includegraphics[scale=0.65]{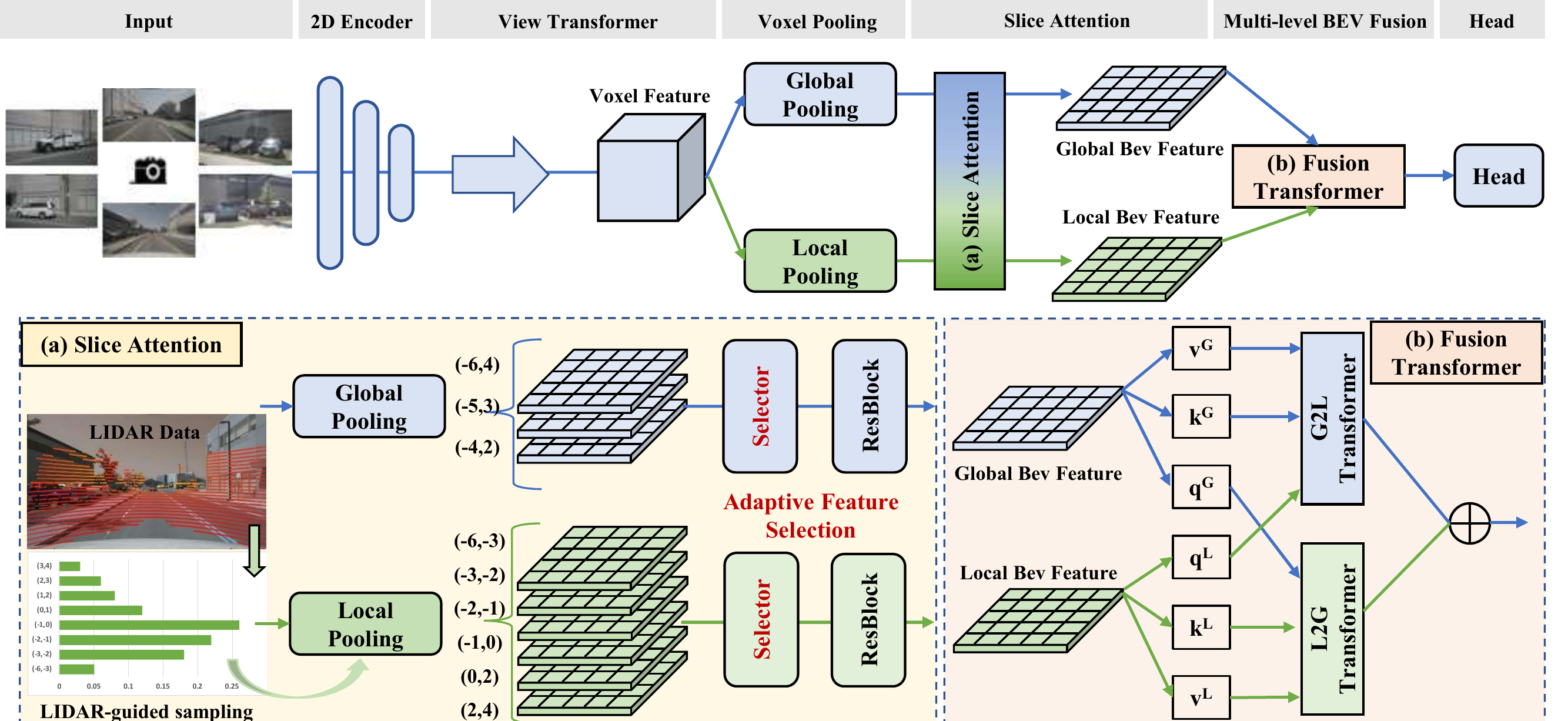}
\caption{The pipeline of the proposed SAN method. Our method constructs the BEV feature based on the global and local slices. We use a two-stage fusion strategy to merge the features of global and local slices for task heads.}
\label{fig:framework}
\end{figure*}

{\bf Multi-View BEV 3D object detection} As a unified representation of the surrounding scene, BEV 3D object detection is becoming prevailing in the multi-view camera systems. Recently, plenty of methods are proposed for multi-view BEV 3D object detection. DETR3D \cite{wang2022detr3d} uses a sparse set of 3D object queries to index the extracted 2D features from multi-view camera images. They make the bounding box prediction per query using the set-to-set loss. BEVDet \cite{huang2021bevdet} first predicts the depth for each camera image and then projects the extracted image features to BEV space by the LSS operation \cite{philion2020lift}. Finally, the task-specific head is constructed upon the BEV feature. BEVDet4D \cite{huang2022bevdet4d} fuses the feature from the previous frame with the current frame to lift the BEVDet paradigm from 3D space to spatial-temporal 4D space. BEVFormer \cite{li2022bevformer} exploits both the spatial and temporal information by interacting with spatial and temporal space through pre-defined grid-shaped BEV queries. PETR \cite{liu2022petr} encodes the position information of 3D coordinates into image features and performs end-to-end object detection based on 3D position-aware features. BEVDepth \cite{li2022bevdepth} reveals that the quality of intermediate depth is the key to improving multi-view 3D object detection. They get explicit depth supervision utilizing encoded intrinsic and extrinsic parameters. PolarDETR \cite{chen2022polar} uses the Polar Parametrization for 3D detection by reformulating position parametrization, velocity decomposition, perception range, label assignment, and loss function in the polar coordinate system. BEVStereo \cite{li2022bevstereo} introduces an effective temporal stereo method to dynamically select the scale of matching candidates for multi-view stereo. They further design an iterative algorithm to update more valuable candidates, making it adaptive to moving candidates. STS \cite{wang2022sts} proposes a surround-view temporal stereo technique to leverage the geometry correspondence between frames across time to improve the quality of depth.

\section{Methods}

Our method follows the pipeline of existing methods such as BEVDepth \cite{li2022bevdepth}, which consist of three components: camera feature extraction, BEV feature construction, and task heads. To be more specific, Given an input multi-view image ${I_k} \in {R^{3 \times H \times W}}$, we adopt a shared backbone model to extract the feature ${F_k} \in {R^{C \times {H_f} \times {W_f}}}$, where k is the index of the camera. we also predict the depth distribution map for each input image ${D_k} \in {R^{D \times {H_f} \times {W_f}}}$. Then we project the camera features to viewing frustum ${V_k} \in {R^{C \times D \times {H_f} \times {W_f}}}$ and construct the flattened BEV feature $B \in {R^{C \times {H_e} \times {W_e}}}$ with the proposed {\bf Slice Attention Module}. Finally, the task-specific heads are applied to the BEV feature. We will first introduce the motivation in Sec. \ref{sec:moti} and then present the proposed Slice Attention Module in Sec. \ref{sec:sam}. The whole framework of our method is illustrated in Fig. \ref{fig:framework}.

\subsection{Motivation}
\label{sec:moti}

 \begin{table}[t]
 	\begin{center}
 		\caption{The mAP results of Traffic Cone, Person and Bus with BEV slices of different height ranges.}
 		\setlength{\tabcolsep}{4mm}{
 			\begin{tabular}{c|ccc}
 				\hline
 				Height & Traffic Cone & Person & Bus  \\
 				\hline
 				\hline 
 				$[-2, 0]$ & 0.087 & 0.0 & 0.001   \\
 				$[ 0, 1]$ & \textbf{0.436} & 0.217 & 0.273   \\
 				$[ 1, 2]$ & 0.367 & 0.245 & 0. 307  \\
 				$[ 2, 3]$ & 0.446 & \textbf{0.265} & 0.340   \\
 				$[ 3, 4]$ & 0.368 & 0.257 & \textbf{0.348}   \\
 				\hline
 		\end{tabular}}
 		\label{tab:motiv}
 	\end{center}
 \end{table}

In the practical applications of autonomous driving, the detection targets vary in shape and size, causing severe bias in visual-based learning. For example, barrier is located at a low height while the truck is located at a high height. However, existing methods like BEVDepth \cite{li2022bevdepth} sum up the frustum features that fall into the same flattened BEV grid. Therefore, they fail to exploit the benefit of different heights for BEV perception. In this section, we intend to demonstrate the motivation for slicing the BEV space based on different heights. We first visualize the heights of annotated 3D bounding boxes according to their object classes. As shown in Fig. \ref{fig:heat_hight_cropped}, different object classes actually have different height distributions. This is consistent with our motivation.

To further study this motivation, we adjust the height range of BEVDepth \cite{li2022bevdepth} and evaluate the 3D object detection performance of different classes as shown in Tab. \ref{tab:motiv}. As can be seen, the traffic cone, which is lower compared with person and bus, shows obviously different performances at different height ranges (0.466 in [-2,0] and 0.368 in [2,4] separately). This indicates that the height range will greatly affect the detection performance of different object classes. This observation inspires us to take better advantage of different heights to improve detection performance. We will introduce the proposed Slice Attention Module in the next section.

\begin{figure}[t]
\includegraphics[scale=0.45]{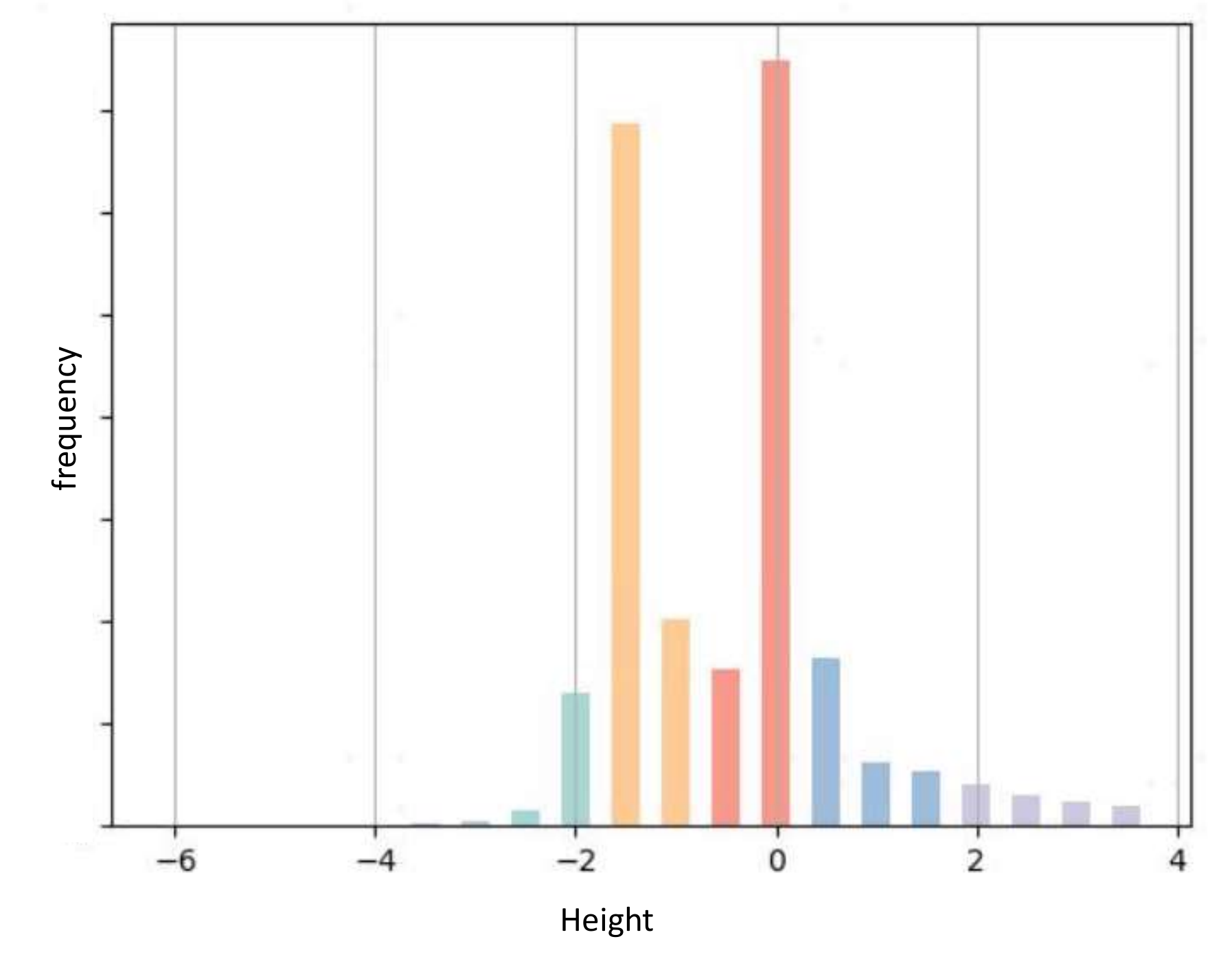}
\centering
\caption{The statistics of LiDAR points along the height dimension. We use this LiDAR histogram to guide the sampling of local slices, which emphasize the informative heights.}
\label{fig:fuse}
\end{figure}

\subsection{Slice Attention Module}
\label{sec:sam}
In this section, we introduce the proposed Slice Attention Module. We define the slice using the height range in BEV space. We will first explain how to sample the BEV space to generate the global and local slices. The global slices are sampled to cover the large height ranges of BEV space. The local slices are sampled to emphasize the informative heights. Then we present our method to fuse the sampled global and local slices with an attention mechanism. Finally, we fuse the global feature and local feature for the task heads.

\subsubsection{Global and Local Slices}

For the multi-view images, we can extract the features by a shared backbone model ${F_k} \in {R^{C \times {H_f} \times {W_f}}}$, where $k$ is the index of the camera. We can aggregate the image features to construct the BEV feature ${B_s} \in {R^{C \times {H_e} \times {W_e}}}$ given the height range $s = [l,u]$ in BEV space. We define a height range as a BEV slice.

\textbf{Global Slices} We empirically determine the global slices as $\{ {s_g}\}  = \left[ {\left[ { - 6,4} \right],\left[ { - 5,3} \right],\left[ { - 4,2} \right]} \right]$. Although the largest range  $[-6,4]$ contains the overall information of the whole space, the corresponding BEV feature representation is significantly different from $[-5,3]$ or  $[-4,2]$. Since the height information is viewed as channel dimension, we adopt a channel-wise attention \cite{hu2018squeeze} to adaptively aggregate the multiple global-level slices. The attention mechanism between three global slices provides a learnable way to fully explore different semantic knowledge and thus improve the global contextual representation in BEV latent space. The attention between three global slices will be necessary to help improve the performance at the global level. We denote the constructed features of global slices as $\{ B_{_{{s_g}}}^i\} $.

\begin{figure}[t]
	\includegraphics[scale=0.63]{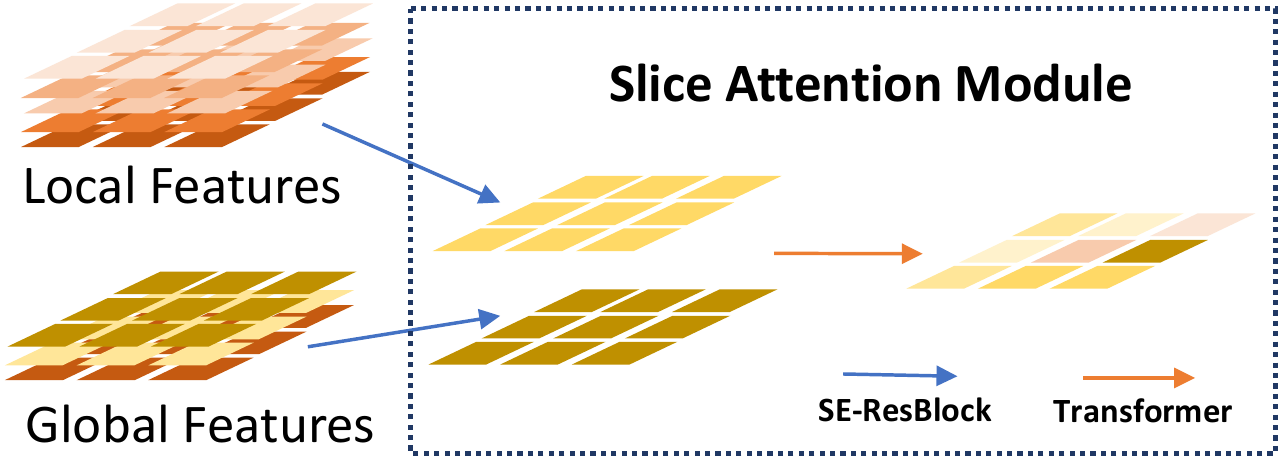}
	\centering
	\caption{The pipeline of slice feature fusion. Our fusion strategy contains two stages. The first stage is based on channel attention to merge local slices and global slices separately. The second stage is based on a dual-branch transformer, which explores the spatial attention.}
	\label{fig:lidar_hist}
\end{figure}

\textbf{Local Slices} The goal of local slices is to emphasize the informative height ranges. We construct the local slices by sampling from the overall range $[-6,4]$. In order to sample reasonable local slices, we present a LiDAR-guided sampling strategy to determine the optimal heights of local slices. We transform the LiDAR points to BEV space and calculate the histogram along the height dimension as shown in Fig. \ref{fig:lidar_hist}. We find that most LiDAR points are located around -2 and 0. However, those regions contain small objects while regions outside [-2,2] contain large objects. In order to sample more effective local slices, we design a novel strategy to consider the distribution differences between classes. Specifically, we accumulate the histogram and choose the local slices from the accumulated distribution. We slice the overall range $[-6,4]$ to six bins, including $[-6,-3]$, $[-3,-2]$, $[-2,-1]$,$[-1,0]$, $[0,2]$, and $[-2,4]$. Similar to global slices, we also utilize the channel attention mechanism to reweight the local slices, which effectively aggregates the information of different heights. The local slices are denoted as $\{ {s_l}\}$ and the aggregated features are denoted as $\{ B_{{s_l}}^j\} $.

\subsubsection{Fusion of Slice Features}
After obtaining the global features $\{ B_{_{{s_g}}}^i\}$ and local features $\{ B_{{s_l}}^j\} $, we can fuse them together into the feature map for task heads. Our method introduces a two-stage attention structure to progressively fuse the features as shown in Fig. \ref{fig:fuse}. In the first stage, we fuse the global features and local features via the attention mechanism. This will generate the global fused feature ${B_g} \in {R^{C \times {H_e} \times {W_e}}}$ and local fused feature ${B_l} \in {R^{C \times {H_e} \times {W_e}}}$. In the second stage, we use a transformer to fuse ${B_g}$ and ${B_l}$ and generate the feature map for task heads.

\begin{figure}[t]
	\includegraphics[scale=0.63]{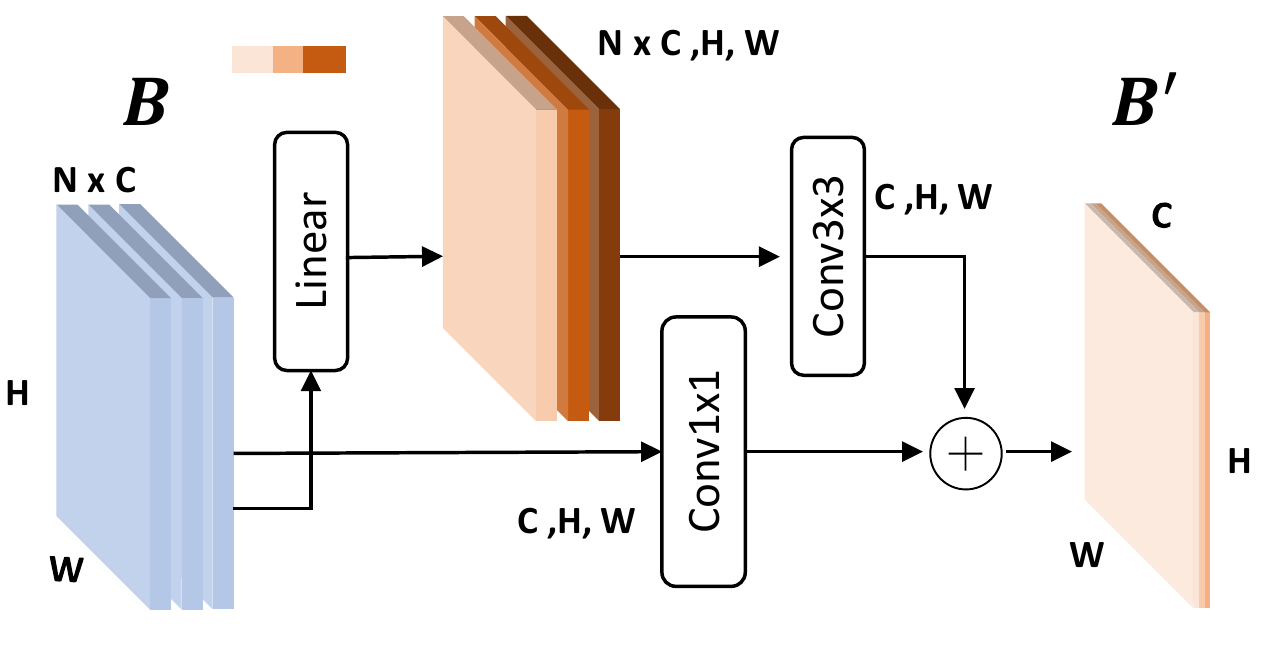}
	\centering
	\caption{Illustration of the SE attention residual block for merging local and global slices separately.}
	\label{fig:fuse1}
\end{figure}

To be more specific, in the first stage, we adopt the attention mechanism similar to the Squeeze-and-Excitation (SE) operation \cite{hu2018squeeze}. Taking local features as an example, the features of local slices are denoted as $\{ B_{_{{s_l}}}^j\}  \in {R^{J \times C \times {H_e} \times {W_e}}}$, where $J$ is the number of local slices. As shown in Fig. \ref{fig:fuse1}, we first use 1x1 convolution to reduce the channel number from ${J \times C}$ to $C$. We use global average pooling to extract the ${J \times C}$ feature and reweight the input feature. Another 3x3 convolution is used to reduce the channel number from ${J \times C}$ to $C$. Finally, we add the two parts to deliver the fused feature ${B_l} \in {R^{C \times {H_e} \times {W_e}}}$. The features of global slices $\{ B_{_{{s_g}}}^i\}$ can be fused into ${B_g}$ in the same way.

In the second stage, we need to fuse ${B_g}$ and ${B_l}$ with a transformer. As shown in Fig. \ref{fig:framework}, the transformer contains two branches (denoted as G2L and L2G) using ${B_g}$ and ${B_l}$ as the inputs. One feature will be transformed into a set of \textit{Key/Value} pairs to interact with features from the other. For example, the \textit{Query/Key/Value} pair in G2L Transformer is: $q=q^L, k=k^G,v=V^G$ where $L$ stands for local-level and $G$ represents global-level. Finally, we sum up the outputs of the two branches to obtain the final feature map for task heads.
\begin{table*}[t]
  \centering
  \caption{3D Object Detection Results on nuScenes val set without CBGS}
    \setlength{\tabcolsep}{1.2mm}{
    \begin{tabular}{c|c|c|c|cccccc}
    \hline
    Method & Voxel Range& Backbone & NDS ↑& mAP ↑ & mATE ↓ & mASE ↓ & mAOE ↓ & mAVE ↓ & mAAE ↓ \\
    \hline\hline
    BEVDepth & [-5,3] & R50 & 0.328 & 0.293 & 0.742 & 0.283 & 0.758 & 1.216 & 0.403 \\
    BEVDepth & [-4,2] & R50 & 0.330 & 0.293 & 0.740 & 0.282 & 0.745 & 1.201 & 0.397 \\
    BEVDepth & [-6,4] & R50 & 0.336 & 0.296 & 0.732 & 0.283 & 0.713 & 1.218 & 0.396 \\
    BEVDet & [-5,3] & R50 & 0.298 & 0.274 & 0.754 & 0.295 & 0.881 & 1.25 & 0.418 \\
    SANet(BEVDet) & slice & R50 & 0.320 & 0.292 & 0.746 & 0.286 & 0.797 & 1.167 & 0.403 \\
    SANet(BEVDepth) & Slice & R50 & 0.366 & 0.310 & 0.705 & 0.278 & 0.608 & 1.070 & 0.300 \\
    \hline
    BEVDepth & [-6,4] & R101 & 0.371 & 0.313 & 0.697 & 0.278 & 0.579 & 1.086 & 0.304 \\
    SANet(BEVDepth) & Slice & R101 & 0.379 & 0.319 & 0.681 & 0.270 & 0.567 & 0.996 & 0.290 \\
    \hline
    \end{tabular}%
    }
  \label{tab:nu_val}%
\end{table*}%

\begin{table*}[t]
	\centering
	\caption{3D Object Detection Results on nuScenes val set with CBGS.}

	\setlength{\tabcolsep}{1.2mm}{
		\begin{tabular}{c|c|c|c|cccccc}
			\hline
			Method & Voxel Range& Backbone & NDS ↑& mAP ↑ & mATE ↓ & mASE ↓ & mAOE ↓ & mAVE ↓ & mAAE ↓ \\
			\hline\hline
			BEVDet & [-5,3] & R50 & 0.372 & 0.299 & 0.724 & 0.273 & 0.578 & 0.929 & 0.266 \\
			PETR & [-5,3] & R50 & 0.381 & 0.313 & 0.768 & 0.278 & 0.564 & 0.923 & 0.225 \\
			BEVDepth & [-5,3] & R50 & 0.470 & 0.341 & 0.619 & 0.273 & 0.451 & 0.462 & 0.198 \\
			SANet(BEVDepth) & Slice & R50 & 0.482 & 0.351 & 0.618 & 0.271 & 0.434 & 0.426 & 0.192 \\
			\hline
		\end{tabular}%
	}
	\label{tab:nu_val_cbgs}%
\end{table*}%

\begin{table*}[!ht]
	\centering
	\caption{3D Object Detection Results of Each Object Class on nuScenes val set.}
	\setlength{\tabcolsep}{2mm}{
		\begin{tabular}{c|ccccccccc}
			\hline
			Method & Truck & trailer & Car  & Bus & Pedestrian & Motorcycle & Bicycle & Barrier & Traffic cone \\
			\hline\hline
			BEVDepth & 0.237 & 0.153 & 0.466 & 0.332 & 0.247 & 0.289 & 0.267 & 0.417 & 0.465 \\
			SANet & 0.244 & 0.165 & 0.491 & 0.358 & 0.265 & 0.302 & 0.272 & 0.432 & 0.503\\
			\hline
			BEVDepth+CBGS &0.269 &0.171& 0.545  & 0.352 & 0.351 & 0.318 & 0.250 & 0.530 & 0.559\\
			SANet+CBGS &0.272&0.166& 0.555  & 0.358 & 0.365 & 0.315 & 0.282 & 0.544 & 0.582\\
			\hline
		\end{tabular}%
	}
	\label{tab:each_cls}%
\end{table*}%
\section{Experiment}
\label{sec:exp}
In this section, we first give the experimental details in Sec. \ref{sec:exp_setup}. Then we evaluate the proposed SAN on nuScenes \cite{caesar2020nuscenes} and compare it with several baseline methods in Sec. \ref{sec:exp_main}. Besides, we also conduct detailed ablation study to evaluate each component of our method in Sec. \ref{sec:exp_abl}. We further show the computational cost in Sec. \ref{sec:exp_cost}.

\subsection{Experimental Details}
\label{sec:exp_setup}
\textbf{Dataset} We use the nuScenes\cite{caesar2020nuscenes} dataset to evaluate the performance of our distillation framework. NuScenes contains 1k sequences, each of which is composed of six groups of surround-view camera images, one group of Lidar data, and their sensor information. The camera images are collected with the resolution of $1600 \times 900$ at 12Hz and the LiDAR frequency for scanning is 20Hz. The dataset provides object annotations every 0.5 seconds, and the annotations include 3D bounding boxes for 10 classes \{Car, Truck, Bus, Trailer, Construction vehicle, Pedestrian, Motorcycle, Bicycle, Barrier,  Traffic cone \}. We follow the official split that uses 750, 150, and 150 sequences as training, validation, and testing sets respectively. So total we get 28130 batches of data for training, 6019 batches for validation, and 6008 batches for testing. 

\textbf{Metrics} We use mean Average Precision(mAP) and Nuscenes Detection Score(NDS) as our main evaluation metrics. We also adopt other officially released metrics concluding Average Translation Error (ATE), Average Scale Error (ASE), Average Orientation Error (AOE), Average Velocity Error (AVE), and Average Attribute Error (AAE). Note that NDS is a weighted sum of mAP and other metric scores.

\textbf{Implementation Details}
We use BEVDepth \cite{li2022bevdepth} as the baseline. The image backbone is ResNet-50 and the input image size is [256,704]. Following BEVDepth, image augmentation includes random cropping, random scaling, random flipping, and random rotation. The BEV feature generated by the model is also augmented by random scaling, random flipping, and random rotation. The base learning rate is 2e-4, and the batch size is 6 for each GPU. During training, we use 8 V100 GPU and the training takes 40 epochs. We decay the learning rate on epochs 23 and 33 with ratio $\alpha=1e-7$. To conduct a fair comparison, all methods share these settings. Apart from BEVDepth, we also evaluate the proposed method on the BEVDet \cite{huang2021bevdet}. 

\begin{figure*}[t]
\includegraphics[width=0.99\textwidth]{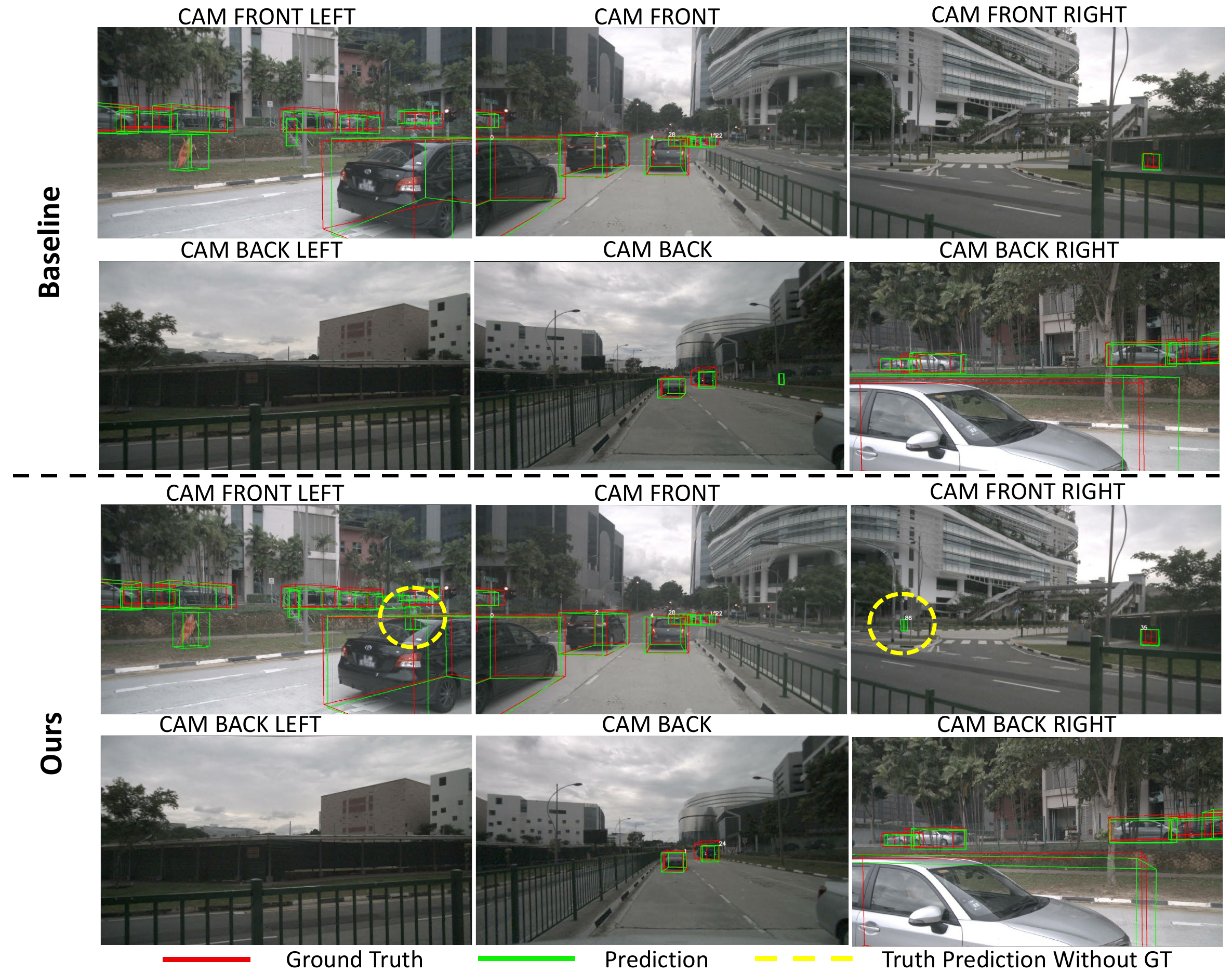}
\centering
\caption{The visualization result of baseline and the SAN. The red box denotes the ground truth, and the green box is the prediction. In this case, our method gives a more accurate prediction, and gives two correct predictions of pedestrians in the yellow circles that do not have labels.}
\label{fig:vis}
\end{figure*}

\subsection{Main Results} 
\label{sec:exp_main}
\textbf{Results on nuScenes val set} We first evaluate our method on nuScenes val set. The baseline methods are BEVDet and BEVDepth. We report the results of BEVDepth under different height ranges [-5,3], [-4,2], and [-6,4]. The default height range of BEVDet and BEVDepth is [-5,3]. As can be seen from Tab. \ref{tab:nu_val}, our method can improve the baseline method by 0.03 in NDS and all the evaluation metrics are also improved. To further evaluate our method, we conduct the experiments with CBGS strategy \cite{zhu2019class}, which will take much longer training time. As can be seen from Tab. \ref{tab:nu_val_cbgs}, our method can still improve performance even with the CBGS strategy. We conduct this experiment based on ResNet-50 in consideration of computation cost.


\textbf{Results of different object classes} Since the motivation of our method is to handle different object classes with different heights. Therefore, we show the results of different object classes in Tab. \ref{tab:each_cls}. We compare the mAP of the proposed SAN and baseline methods. For the results without CBGS strategy, SAN outperforms the baseline BEVDepth in each object class. The performance gain of traffic cone even reaches 0.038. For the results with CBGS, the SAN also shows significant improvement. For example, our method improves the baseline BEVDepth by 0.032 in bicycles and 0.023 in traffic cones. These results show that our method gives different attention to objects with different shapes.

\begin{figure*}[t]
\includegraphics[width=1\textwidth]{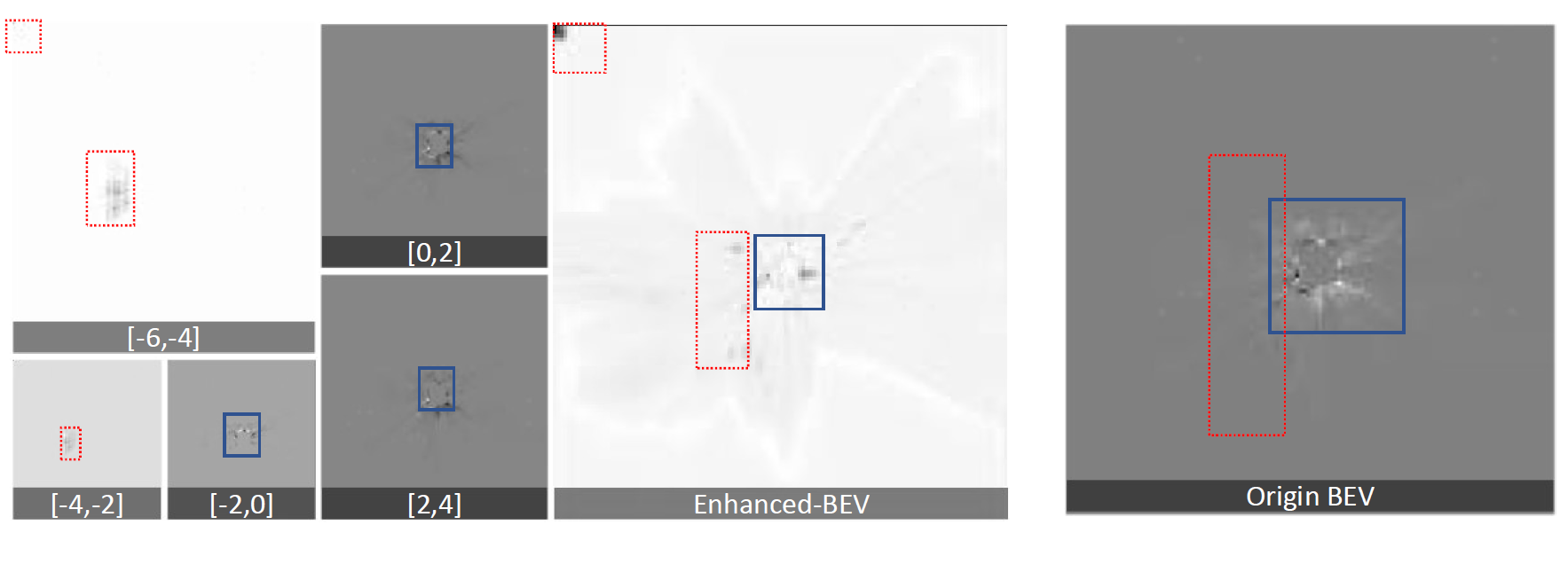}
\centering
\vspace{-0.9cm}
\caption{The visualization result of the baseline BEV feature and SAN BEV feature. As can be seen, the features of different slices can capture different objects. For example, the original feature fails to capture the top-left object, while our enhanced feature successfully capture this object.}
\label{fig:bevvis}
\end{figure*}

\textbf{Qualitative results} We show the qualitative results of the baselines and our method. As can be seen from Fig. \ref{fig:vis}, the proposed SAN improves the performance of 3D object detection. In this figure, we compare the results of SAN and BEVDepth \cite{li2022bevdepth}. We also show the feature visualization in Fig. \ref{fig:bevvis}. As can be seen from this figure, the original BEV feature does not capture the top left object, while our method fuses the features of different slices. Therefore, the enhanced BEV feature successfully captures the top left object.

\begin{table}[t]
	\centering
	\caption{Ablation Study of Global and Local Slices.} 
	\setlength{\tabcolsep}{5mm}{
		\begin{tabular}{c|c|cc}
			\hline
			Local & Global & NDS & mAP  \\
			\hline\hline
			. & . & 0.330 & 0.296  \\
			\checkmark & .  & 0.351  & 0.310 \\
			. & \checkmark & 0.343  & 0.307 \\
			\checkmark & \checkmark  & \textbf{0.366} & \textbf{0.310} \\
			\hline
		\end{tabular}%
	}
	\label{tab:global_local}%
\end{table}%

\subsection{Ablation study}
\label{sec:exp_abl}
\textbf{Global and Local Slices} Our method uses both the global and local slices to construct the BEV feature. The global slices aim to cover the large ranges of BEV height while the local slices aim to emphasize the informative heights. Therefore, we conduct an ablation study to evaluate the contributions of global and local slices. As shown in Tab. \ref{tab:global_local}, both types contribute to performance improvement. 

\textbf{LiDAR-Guided Sampling}
In this paper, we propose to use LiDAR-guided sampling strategy to obtain the local slices. Therefore, we conduct the ablation study to evaluate the contribution of this component. For a fair comparison, we all use the global slices. As can be seen from Tab. \ref{tab:slice}, the LiDAR-guided sampling strategy can improve the NDS of average local sampling by 0.07, demonstrating the effectiveness of the proposed sampling strategy.
\begin{table}[t]
	\centering
	\caption{Ablation Study of LiDAR-Guided Sampling.} 
	\setlength{\tabcolsep}{6mm}{
		\begin{tabular}{c|cc}
			\hline
			Statistics Local  & NDS & mAP \\
			\hline\hline
			.  & 0.359  & 0.310 \\
			\checkmark  & \textbf{0.366} & \textbf{0.310} \\
			\hline
		\end{tabular}%
	}
	\label{tab:slice}%
\end{table}%

\begin{table}[t]
	\centering
	\caption{Ablation Study of Fusion Strategy.}
	\vspace{-0.2cm}
	\setlength{\tabcolsep}{2.5mm}{
		\begin{tabular}{c|c|c|c}
			\hline
			Method & Voxel Range& NDS ↑& mAP ↑ \\
			\hline\hline
			SA-Mean & local Only & 0.332 & 0.296  \\
			SA-SE & local Only & 0.350 & 0.298  \\
			SA-SE-Mean & local + Global & 0.359 & 0.311  \\
			SA-SE-SE & local + Global & 0.361 & 0.310  \\
			SA-SE-Trans & local + Global & 0.366 & 0.310  \\
			\hline
		\end{tabular}%
	}
	\label{tab:fusion}%
\end{table}%

\begin{table*}[t]
	\begin{center}
 		\caption{Computational cost. We compare the proposed SANet with the baseline method BEVDepth with ResNet-50 and ResNet-101 as the backbones. As can be seen, our method will introduce some additional computational cost. However, this is because we simply repeat the LSS operation many times to generate the features of slices. Careful engineering optimization can significantly improve the efficiency.}
		\setlength{\tabcolsep}{2mm}{
			\begin{tabular}{c|c|c|cc|ccc}
				\hline
				Method & Backbone & NDS & FPS & Model Size(MB)& Image backbone(ms) & pooling(ms) & Fusion(ms)    \\
				\hline
				\hline
				SANet & R50 & 0.366 & 15.4 & 911.0 & 0.53 &  23.12 &  0.50 \\
				SANet & R101 & 0.379 & 14.3 & 1128.7 & 0.55 & 26.22 & 0.45  \\
				BEVDepth & R50 & 0.330 & 24.3 & 870.0 & 0.54 & 26.23 & . \\
				BEVDepth & R101 & 0.371 & 19.6 & 1087.1 & 0.55 & 26.26 & .  \\
				\hline
		\end{tabular}}
		\label{tab:fps}
	\end{center}
\end{table*}
\textbf{Fusion Strategy}
The fusion strategy also plays an important role in merging the local and global slices. In short, our fusion strategy contains two stages. The first stage merges the local and global slices respectively. The second stage fuses the merged local and global features for task heads. In this part, we evaluate the fusion strategy based on BEVDepth with ResNet-50. Mean denotes adding the BEV features together. SE denotes the Squeeze-and-Excitation Attention residual block. Trans means the designed two branches transformer. As can be seen in Tab. \ref{tab:fusion}. Using SE in the first stage and Trans in the second stage achieves the best performance compared with the alternatives. Nevertheless, all the fusion strategies can achieve considerable improvements compared with the baseline, demonstrating the effectiveness of the proposed SAN.
\subsection{Computational Cost}
\label{sec:exp_cost}
In this section, we report the computational cost of SAN. As shown in Tab. \ref{tab:fps}, our method introduces additional computational and storage cost to the baseline methods. To be more specific, when the backbone is ResNet-101, our method introduces 41 MB storage cost and 27\% slower than the BEVDepth baseline. The most time-consuming step is building the features of global and local slices. However, this is because our current implementation simply repeats the LSS \cite{philion2020lift} operations. More careful engineering optimization can help to reduce the computational cost of SAN, which will be our future work. 
\section{Limitation}
Although the proposed SAN is simple yet effective, our method still has some limitations. One limitation is the additional computational and storage cost as mentioned above. However, we believe careful engineering optimization can solve this problem. Besides, our method follows the BEVDepth \cite{li2022bevdepth} pipeline, which is sensitive to the accuracy of depth values or the depth distributions. How to apply SAN to baseline methods such as BEVFormer \cite{li2022bevformer} is still a problem, which will also be our future work.
\section{Conclusion}
In summary, we propose a novel method named Slice Attention Network for BEV 3D object detection in this paper. Instead of summing up the frustum features that fall into the same flattened BEV grid, our method explores the benefit of different heights in BEV space. We extract the BEV features of global and local slices. The global slices aim at covering the large height ranges while the local slices aims at emphasizing informative local height ranges. To improve the performance, we propose to sample the local slices based on the histogram of LiDAR points along the height dimension. The features of local and global slices are fused by a two-stage strategy for task heads. We use BEVDepth as the baseline method and conduct detailed experiments to demonstrate the effectiveness of BEV-SAN.
\begin{table*}[t]
  \centering
  \caption{Comparisons of Generalization ability with different methods on the validation set of unseen environment \cite{caesar2020nuscenes}. The unseen environment includes night-time and rainy data. All methods utilize ResNet 50 \cite{he2016deep} as backbone. } 
    \setlength{\tabcolsep}{0.8mm}{
    \begin{tabular}{c|c|c|c|cccccc}
    \hline
      Test on&Method & Backbone & \cellcolor{lightgray} NDS ↑& \cellcolor{lightgray} mAP ↑ & mATE ↓ & mASE ↓ & mAOE ↓ & mAVE ↓ & mAAE ↓ \\
    \hline\hline
     \multirow{2}{*}{Night} &BEVDepth\cite{li2022bevdepth}  & R50 & 0.170 & 0.124 & 0.847 & 0.463 & 0.906 & 1.855 & 0.696 \\
   &BEV-SAN & R50 & \textbf{0.210} & \textbf{0.129} & \textbf{0.827} & \textbf{0.466} & \textbf{0.670} & \textbf{1.655} & \textbf{0.584} \\
    \hline
    \multirow{2}{*}{Rainy} &BEVDepth\cite{li2022bevdepth}  & R50 & 0.363 & 0.305 & 0.722 & 0.298 & 0.662 & 0.915 & 0.289 \\
   &BEV-SAN & R50 & \textbf{0.396} & \textbf{0.314} & \textbf{0.711} & \textbf{0.296} & \textbf{0.629} & \textbf{0.664} & \textbf{0.242} \\
    \hline
    \end{tabular}%
    }
  \label{tab:gen}%
\end{table*}%

\begin{table*}[t]
  \centering
  \caption{Comparisons of the Robustness ability with different methods on the validation set \cite{caesar2020nuscenes}. We design a special experiment setting in which one camera breaks down or is occluded. And we occlude the front-view images in inference time.} 
    \setlength{\tabcolsep}{6.1mm}{
    \begin{tabular}{c|c|c|c|c}
    \hline
    Occlude & Method & Backbone & \cellcolor{lightgray} NDS ↑& \cellcolor{lightgray} mAP ↑  \\
    \hline\hline
    & BEVDepth\cite{li2022bevdepth}  & R50 & 0.336 & 0.296  \\
    \hline
    \multirow{2}{*}{Front}  & BEVDepth\cite{li2022bevdepth}  & R50 & 0.318 & 0.228  \\
     & Ours(BEVDepth) & R50 & \textbf{0.325} & \textbf{0.258}  \\
     \hline
      \multirow{2}{*}{Front-Left}  & BEVDepth\cite{li2022bevdepth}  & R50 & 0.331 & 0.265  \\
     & Ours(BEVDepth) & R50 & \textbf{0.332} & \textbf{0.279}  \\
     \hline
      \multirow{2}{*}{Front-Right}  & BEVDepth\cite{li2022bevdepth}  & R50 & 0.326 & 0.242  \\
     & Ours(BEVDepth) & R50 & \textbf{0.330} & \textbf{0.271}  \\
     \hline
    \end{tabular}%
    }
  \label{tab:rob}%
\end{table*}%
\section{Appendix}
In the supplementary material, we first present additional related work of transformer network in Sec .\ref{sec:1} since we utilize dual-branch transformer module to fuse the global and local slices. In Sec .\ref{sec:2}, we then provide additional and detailed cross domain training strategy. In Sec .\ref{sec:3}, we explore the generalization ability of our proposed BEV-SAN by evaluating the performance on unseen and challenging data distribution. In Sec .\ref{sec:4}, we demonstrate the robustness of our method by comparing with baseline methods when encountering cameras malfunctioning. 


\subsection{Additional related works}
\label{sec:1}
\textbf{Vision transformer.}
Transformer network was first introduced for neural machine translation tasks \cite{vaswani2017attention}, and the encoder and decoder of transformer leverage self-attention mechanism to extract better feature representation and reserve contextual information \cite{vaswani2017attention,parikh2016decomposable,lin2017structured}. 
Vision Transformer (ViT) \cite{dosovitskiy2020image,touvron2021training} first brings a transferring in backbone architectures for computer vision, which is transferred from CNNs to Transformers.
This seminal work has led to subsequent research that aims to improve its utility \cite{liu2022video}. 
Meanwhile, Swin Transformer \cite{liu2021swin} is a practical backbone for various image recognition tasks, which adopts the inductive biases of locality, hierarchy and translation invariance. DeiT \cite{touvron2021training} focuses on improving the efficiency and practicality of transformer network, it proposes several training strategies that allows ViT to be effective when training on smaller image datasets. In this paper, we introduce a dual branches transformer block to fuse global an local-level BEV slices and generate the fused BEV feature map for task heads.

\subsection{Additional implementation details}
\label{sec:2}
Our training process can be regarded as an end-to-end training. Firstly, in order to fully leverage the feature extraction ability of the model~\cite{li2022bevdepth}, we load the backbone of ImageNet pretrained parameters. Then we train the model with slice-attention module for 28 epochs with \textbf{CBGS}~\cite{zhu2019class} and 40 epochs without. It should be noted that we freeze the backbone starting from epoch 23 and fine-tune the slice-attention module and detection head in the rest of the epochs. We adopt $256\times 704$ as image input size and the same data augmentation methods as \cite{li2022bevdepth}. We apply AdamW \cite{loshchilov2017decoupled} optimizer with 2e-4 learning rate. We decay the learning rate on epochs 19, 23, and 33 with ratio $\alpha=1e-7$. As for further detailed image augmentation process, we follow BEVDepth and adopt random cropping, random scaling, random flipping, and random rotation. The BEV feature generated by the model is also augmented by random scaling, random flipping, and random rotation. All experiments are conducted on NVIDIA Tesla V100 GPUs. 

\subsection{Additional generalization exploration}
\label{sec:3}

Slice-attention module leverages the attention mechanism of Transformer to fuse the features from different global information to construct a more comprehensive BEV feature. Therefore, BEV-SAN is of better generalization ability in more display scenarios after integrating multiple levels of information. We conduct further experiments on some particular scenarios like rainy and night in NuSences dataset to demonstrate the superiority generalization ability of BEV-SAN.

As shown in Tab. \ref{tab:gen}, the baseline can only achieve 0.170 and 0.124 in NDS and mAP, respectively on the night validation set. Due to the faint light condition at night, the camera based method will encounter great challenges. However, we observe that BEV-SAN shows satisfying performance under such severe condition with 0.210 NDS and 0.129 mAP, respectively. As for rainy validation set, we notice that BEV-SAN also outperforms the baseline with significant margin by over 3\% in NDS. These results verify the generalization ability of BEV-SAN.

\subsection{Additional robustness exploration}
\label{sec:4}
Though there are lots of recent works on autonomous driving systems, only a few of them \cite{philion2020lift,li2022bevformer} explore the robustness of the proposed methods. LSS \cite{philion2020lift} presents the performance under extrinsic noises and camera dropout at test time. Following previous work, we aim to give a qualitative analysis of our method under camera missing condition. 
Camera image missing occurs when one camera breaks down or is occluded. Multi-view images provide panoramic visual information, yet it can also face the condition when one of them is absent in the real-world. Therefore, it is necessary to evaluate the robustness of our method when encountering camera view missing. 

As shown in Tab. \ref{tab:rob}, among six
cameras of nuScenes dataset, front-view data are the most important, and their absence leads
to a drop of 1.8\% NDS and 6.8\% mAP on BEVDepth \cite{li2022bevdepth}. In term of our proposed method, front-view camera missing only leads to a drop of 1.1\% NDS and 3.8\% mAP, which demonstrates that BEV-SAN has a great potential on robustness. For other view missing, the results show similar tendency.
{\small
\bibliographystyle{ieee_fullname}
\bibliography{egbib}
}

\end{document}